\documentclass{article}

\usepackage{graphicx}
\usepackage{url}
\usepackage{latexsym}
\usepackage{bm}
\usepackage{amsmath}
\usepackage{amssymb}
\usepackage{amsthm}
\usepackage{amsfonts}
\usepackage{algorithm}
\usepackage{algorithmic}
\usepackage{epstopdf}
\usepackage{float}
\usepackage{caption}
\usepackage{epsfig}
\usepackage{graphics}
\usepackage{subfigure}
\usepackage{psfrag}
\usepackage{cite}

\begin{document}

\title{SLDR-DL: A Framework for SLD-Resolution with Deep Learning}

\author{Cheng-Hao Cai \\
Xiamen Deep-Logic Laboratory, China. \\
Email: chenghao.cai@outlook.com}

\date{}

\maketitle

\begin{abstract}
This paper introduces an SLD-resolution technique based on deep learning. This technique enables neural networks to learn from old and successful resolution processes and to use learnt experiences to guide new resolution processes. An implementation of this technique is named SLDR-DL. It includes a Prolog library of deep feedforward neural networks and some essential functions of resolution. In the SLDR-DL framework, users can define logical rules in the form of definite clauses and teach neural networks to use the rules in reasoning processes.
\end{abstract}


\section{Introduction}	
\label{sec:intro}

SLDR-DL is a general purpose framework for SLD-resolution with deep learning. The name ``SLD-resolution" is the abbreviation of SL-resolution for definite clauses \cite{DBLP:conf/ifip/Kowalski74,DBLP:journals/jacm/AptE82}, while the name ``SL-resolution" is the abbreviation of linear resolution with selection function \cite{DBLP:journals/ai/KowalskiK71}. In the SLDR-DL framework, computers can reason and learn to reason by using definite clauses \cite{DBLP:journals/jsyml/Horn51} and deep neural networks \cite{LeCun15}. The core concept of this framework is to train neural networks via successful resolution processes and to use the trained neural networks to guide new resolution processes heuristically.

The SLDR-DL framework has two aims: The first is to simulate the interaction between learning and reasoning: Systems are expected to learn from reasoning processes and to use learnt experiences to guide new reasoning processes. The second is to solve the problem of combinatorial explosion in automated reasoning \cite{Bundy1983The}: When a problem becomes complex, its search tree of reasoning often grows rapidly. Many complex problems fail to be resolved because it is difficult to find true answers from huge search trees.

The SLDR-DL framework is implemented in SWI-Prolog \cite{DBLP:journals/tplp/WielemakerSTL12}. Its source code can be downloaded from GitHub\footnote{The source code can be downloaded from https://github.com/cchrewrite/SLDR-DL/}, and it provides:
\begin{itemize}
\item A Prolog library of deep neural networks.
\item An implementation of SLD-resolution with deep learning.
\item Some worked examples.
\end{itemize}
The source code will be updated continuously.

This paper consists of the following sections: Section \ref{sec:rworks} introduces related works briefly. Section \ref{sec:sld} introduces the theory behind SLDR-DL. Section \ref{sec:guide} provides a practical guide about how to use this framework. Section \ref{sec:summ} provides a summary of this paper.

\section{Related Works}
\label{sec:rworks}

SLD-resolution \cite{DBLP:conf/ifip/Kowalski74} is a fundamental technique of automated reasoning. It has been used in many fields of artificial intelligence. For instance, Prolog is a programming language based on this technique \cite{DBLP:conf/hopl/ColmerauerR93,DBLP:journals/tplp/WielemakerSTL12}. Mathematical reasoning processes, such as pattern matching, variable substitution and implication, can be simulated in Prolog \cite{Bundy1983The}. Also, belief-desire-intention (BDI) agents can be developed with this technique \cite{DBLP:conf/maamaw/Rao96,Bordini2007Programming}.

Recently, many researchers have explored how to use deep learning to realise reasoning: For instance, Irving et al. \cite{DBLP:conf/nips/IrvingSAECU16} have developed DeepMath which uses deep neural networks to select possible premises in automated theorem proving processes. Also, Serafini and Garcez \cite{DBLP:journals/corr/SerafiniG16} have proposed Real Logic for the integration of learning and reasoning. In the field of reinforcement, Garnelo et al. \cite{DBLP:journals/corr/GarneloAS16} have tried to teach deep neural networks to generate symbols and build representations. In addition, Cai et al. \cite{DBLP:journals/corr/CaiKXS17} have explored the possibility of using deep feedforward neural networks to guide algebraic reasoning processes.

\section{SLD-Resolution with Deep Learning}
\label{sec:sld}

SLD-resolution with deep learning is a fundamental technique of the SLDR-DL framework. It enables deep neural networks to guide new resolution processes after learning from old and successful resolution processes.

\subsection{SLD-Resolution}
\label{subsec:sld}

SLD-Resolution \cite{DBLP:conf/ifip/Kowalski74} is a process deciding whether a goal is satisfiable with a set of definite clauses. It is based on unification, definite clauses and resolution. In this section, we assume that readers have been familiar with these techniques, and only essential definitions and simple examples are carried out to aid the readability.

\subsubsection{Unification}
\label{subsubsec:uni}

Unification \cite{DBLP:books/el/RV01/BaaderS01} is one of the core algorithms in logic programming. It can make two terms become equivalent ones by substitution:

\textbf{Definition 1 (Term).} A term is a constant, a variable or a functor followed by a sequence of terms. Formally, it is defined as:
\begin{equation}
\label{def_term}
t := c~|~v~|~(f,t_1,t_2,\cdots,t_m)
\end{equation}
where $ c $ is a constant, $ v $ is a variable, $ f $ is a functor and $ t_1,t_2,\cdots,t_m $ are terms.

A term can be used to represent facts. For instance, if $ (Love,x,y) $ means ``$ x $ loves $ y $", and $ (Know,p,q) $ means ``$ p $ knows $ q $", then ``Haibara knows that Conan loves Ran" can be represented as $ (Know,Haibara,(Love,Conan,Ran)) $.

\textbf{Definition 2 (Unification).} Unification is a process deciding whether or not two terms can be equivalent ones by substituting their variables for other variables or constants. The standard unification algorithm usually unifies two terms by computing the most general unifier (MGU). A unifier of two terms is a set of substitutions which can make the two terms to be equivalent ones, and the MGU of the two terms is the unifier which can be unified with all unifiers of the two terms. Formally, unification produces the MGU $ \phi $ of two terms $ t_a $ and $ t_b $ such that:
\begin{equation}
\label{def_unification}
t_a[\phi] \equiv t_b[\phi]
\end{equation}

For instance, $ (Know,p, (Love,x,Ran)) $ and $ (Know,Haibara,q) $ can be unified by applying the MGU $ \{p/Haibara, q/(Love,x,Ran)\} $, where ``$ / $" is the substitution operation.

\subsubsection{Definite Clauses}
\label{subsubsec:claf}

Definite clauses \cite{DBLP:journals/jsyml/Horn51} are used to represent relations between terms, especially their implication relations:

\textbf{Definition 3 (Definite Clause).} A definite clause is an implication relation between multiple premises and a single conclusion. Formally, it is defined as:
\begin{equation}
\label{def_clause}
p_1 \land p_2 \land \cdots \land p_n \implies q
\end{equation}
where $ p_1,p_2,\cdots,p_n $ are premises, ``$ \land $" is the logical AND, ``$ \implies $" is the implication symbol, and $ q $ is a conclusion. All the premises and the conclusion are terms.

\textbf{Definition 4 (Disjunction Form).} The disjunction form of the definite clause $ p_1 \land p_2 \land \cdots \land p_n \implies q $ is:
\begin{equation}
\label{def_disjform}
q \lor \lnot p_1 \lor \lnot p_2 \lor \cdots \lor \lnot p_n
\end{equation}
where ``$ \lor $" is the logical OR, and ``$ \lnot $" is the logical NOT. In this formula, $ q $ is called a positive literal, and $ \lnot p_1,\lnot p_2, \cdots,\lnot p_n $ are called negative literals. Formula (\ref{def_clause}) and Formula (\ref{def_disjform}) can be proved to be logically equivalent \cite{DBLP:books/daglib/0017977}.

For instance, ``if $ a $ is bigger than $ b $, and $ b $ is bigger than $ c $, then $ a $ is bigger than $ c $" can be represented as: $ (Bigger,a,b) \land (Bigger,b,c) \implies (Bigger,a,c) $, and its disjunction form is $ (Bigger,a,c) \lor \lnot (Bigger,a,b) \lor \lnot  (Bigger,b,c) $.

\subsubsection{Resolution}
\label{subsubsec:resolution}

The resolution algorithm \cite{DBLP:conf/ifip/Kowalski74} can decide whether or not a goal is satisfiable:

\textbf{Definition 5 (Goal).} A goal is a definite clause with an empty conclusion $ g_1 \land g_2 \land \cdots \land g_n \implies $, and its disjunction form is $ \lnot g_1 \lor \lnot g_2 \lor \cdots \lor \lnot g_n $.

\textbf{Definition 6 (Rule).} A rule is a definite clause with a conclusion $ p_1 \land p_2 \land \cdots \land p_n \implies q $, and its disjunction form is $ q \lor \lnot p_1 \lor \lnot p_2 \lor \cdots \lor \lnot p_n $. In particular, a rule is called an assertion if its premise is empty. In this case, it becomes $ \implies q $, and its disjunction form is $ q $.

\textbf{Definition 7 (SLD-Resolution).} SLD-resolution is a process analysing goals by applying rules: Assume that a goal is $ \lnot g_1 \lor \lnot g_2 \lor \cdots \lor \lnot g_i \lor \cdots \lor \lnot g_n $ and a rule is $ q \lor \lnot p_1 \lor \lnot p_2 \lor \cdots \lor \lnot p_n $. Firstly, a negative literal $ \lnot g_i $ is selected from the goal. Secondly, unification is used to compute the MGU $ \phi $ such that $ g_i[\phi] \equiv q[\phi] $. Lastly, if the unification process is successful, then $ \lnot g_i $ is replaced by $ \lnot p_1 \lor \lnot p_2 \lor \cdots \lor \lnot p_n $, and the goal becomes $ \lnot g_1[\phi] \lor \lnot g_2[\phi] \lor \cdots \lor \lnot p_1[\phi] \lor \lnot p_2[\phi] \lor \cdots \lor \lnot p_n[\phi] \lor \cdots \lor \lnot g_n[\phi] $. In particular, if the rule is an assertion, then the goal becomes $ \lnot g_1[\phi] \lor \lnot g_2[\phi] \lor \cdots \lor \lnot g_{i-1}[\phi] \lor \lnot g_{i+1}[\phi] \lor \cdots \lor \lnot g_n[\phi] $, as $ \lnot g_i $ is eliminated. The above process is run iteratively until the goal is empty, and backtracking is used to select new rules when unification fails.

For instance, given three rules:
\begin{equation}
\label{rule_exp1}
(Bigger,4,2)
\end{equation}
\begin{equation}
\label{rule_exp2}
(Bigger,2,1)
\end{equation}
\begin{equation}
\label{rule_exp3}
(Bigger,a,c) \lor \lnot (Bigger,a,b) \lor \lnot (Bigger,b,c)
\end{equation}
and a goal:
\begin{equation}
\label{goal_exp}
\lnot (Bigger,x,y)
\end{equation}
SLD-resolution can prove $ (Bigger,4,2) $, $ (Bigger,2,1) $ and $ (Bigger,4,1) $ by resolving the goal (``$ \square $" is used to represent ``empty"):
\begin{equation}
\label{sld_exp1}
\lnot(Bigger,x,y) \xrightarrow{Rule~(\ref{rule_exp1})} \square[x/4,y/2]
\end{equation}
\begin{equation}
\label{sld_exp2}
\lnot(Bigger,x,y) \xrightarrow{Rule~(\ref{rule_exp2})}
\square[x/2,y/1]
\end{equation}
or
\begin{equation}
\label{sld_exp3}
\begin{aligned}
& \lnot(Bigger,x,y) \\
\xrightarrow{Rule~(\ref{rule_exp3})}~ & \lnot (Bigger,x,b)[a/x,c/y] \lor \lnot (Bigger,b,y)[a/x,c/y] \\
\xrightarrow{Rule~(\ref{rule_exp1})}~ & \lnot (Bigger,2,y)[a/x,b/2,c/y,x/4,] \\
\xrightarrow{Rule~(\ref{rule_exp2})}~ & \square[a/x,b/2,c/y,x/4,y/1]
\end{aligned}
\end{equation}

\subsection{Deep Neural Networks}
\label{subsec:dnn}

Deep neural networks are used to select rules during the process of SLD-resolution. In this section, we assume that readers have been familiar with deep neural networks, and only essential definitions are carried out.

\textbf{Definition 8 (Deep Feedforward Neural Network).} A deep feedforward neural network (DFNN) \cite{LeCun15} is a neural network satisfying: (1) It has 5 or more than 5 hidden layers. (2) Two neighbouring layers are fully connected. (3) It does not have any recurrent connections. A DFNN can map an input vector to an output vector.

\textbf{Definition 9 (Back-Propagation).} Back-propagation \cite{DBLP:journals/nn/Hecht-Nielsen88a} is a supervised learning method of neural networks. Given an input vector, a feedforward neural network can map it to an output vector, compute an error between the output vector and a target vector and use back-propagation to transfer the error to different layers and update the neural network.

\subsection{The SLDR-DL Framework}
\label{subsec:sldrdl}

The SLDR-DL framework is the combination of SLD-resolution and DFNNs. It enables the deep neural networks to guide and learn to guide resolution processes.

\subsubsection{The Framework Structure}
\label{subsubsec:struc}

The core part of the SLDR-DL framework is an implementation of SLD-resolution with DFNNs.

\textbf{Definition 10 (SLD-Resolution with DFNNs).} SLD-resolution with DFNNs is adapted from the standard SLD-resolution (see \cite{DBLP:conf/ifip/Kowalski74} and Definition 7). When resolving a goal, the following strategy is used: Firstly, a goal literal is encoded to an input vector. Secondly, a trained neural network is used to maps the input vector to an output vector. Thirdly, the output vector is decoded to a ranking list of rules. Finally, rules are applied to the goal according to the ranking list. The methods of encoding and decoding will be discussed in \ref{subsubsec:encoding_decoding}.

In the above process, the neural network is used to predict the ranking list of rules for the given literal. Therefore, the neural network must learn to rank the rules before it is used for prediction.

\textbf{Definition 11 (Learning by SLD-Resolution).} Learning by SLD-resolution is a technique which trains neural networks by using successful resolution processes. Before learning, a goal must be successfully resolved, and records of resolution must be produced. Each record consists of a selected literal and the name of a rule which has been applied to the literal. These records are used to train the neural network: Firstly, the selected literal is encoded to an input vector. Then the name of the rule is encoded to a target vector. Finally, the input vector and the target vector are used to train the neural network with the back-propagation algorithm \cite{DBLP:journals/nn/Hecht-Nielsen88a}. The methods of encoding and decoding will be discussed in Section \ref{subsubsec:encoding_decoding}.

Based on the resolution function and the learning function discussed above, an SLDR-DL system usually consists of:
\begin{itemize}
\item A deep neural network.
\item A rule set for resolution and the encoding and decoding of rules.
\item A symbol set for the encoding of literals.
\end{itemize}

\textbf{Definition 12 (Rule Set).} A rule set contains logical rules with unique names and unique IDs. These rules are definite clauses written in disjunction form. Their IDs should be positive integers.

For instance, a rule set can be:
\begin{table}[H]
\centering
\begin{tabular}{|c|c|c|}
\hline
ID & Name & Rule \\
\hline
\hline
1 & Bigger42 & $ (Bigger,4,2) $ \\
\hline
2 & Bigger21 & $ (Bigger,2,1) $ \\
\hline
3 & BiggerABC & $ (Bigger,a,c) \lor \lnot (Bigger,a,b) \lor \lnot (Bigger,b,c) $ \\
\hline
\end{tabular}
\end{table}

\textbf{Definition 13 (Symbol Set).} A symbol set contains symbols with unique IDs. The IDs should be positive integers.

For instance, a symbol set can be:
\begin{table}[H]
\centering
\begin{tabular}{|c|c|}
\hline
ID & Symbol \\
\hline
\hline
1 & $ Vble $ \\
\hline
2 & $ Bigger $ \\
\hline
3 & $ 1 $ \\
\hline
4 & $ 2 $ \\
\hline
5 & $ 4 $ \\
\hline
\end{tabular}
\end{table}

\subsubsection{Encoding and Decoding}
\label{subsubsec:encoding_decoding}

To enable neural networks to guide resolution processes, encoding and decoding are required, as discussed by Section \ref{subsubsec:struc}: (1) Selected literals should be encoded to input vectors; (2) Rules should be encoded to target vectors; (3) Output vectors should be decoded to ranking lists of rules. In the SLDR-DL framework, we have implemented the following encoding or decoding methods:
\begin{itemize}
\item Given a symbol set $ s $, a predefined depth $ d $ and a predefined breadth $ b $, a negative literal $ \lnot l $ is encoded to a vector via the following steps: Firstly, all variables of $ l $ are replaced by a notation ``$ Vble $". Let $ l_{NV} $ denote this new expression. Secondly, $ l_{NV} $ is rewritten to a completed term $ l_{Comp} $ with the depth $ d $ and the breadth $ b $. All positions exceed the depth and the breadth are omitted, and empty positions are filled by a notation ``$ Empty $". Thirdly, $ l_{Comp} $ is flatten to a list $ l_{List} $. Finally, $ l_{List} $ is represented as a vector by using the one-hot encoding \cite{DBLP:conf/acl/TurianRB10}. Activated positions of the one-hot encoding are decided by the IDs of symbols in $ s $. In particular, ``$ Empty $" is encoded to a zero block.
\item A rule is encoded to a vector via the one-hot encoding \cite{DBLP:conf/acl/TurianRB10}, according to its unique ID in a rule set.
\item An output vector $ (y_1,y_2,\cdots,y_m) $ is decoded to a ranking list via the following steps: Firstly, IDs are attached to all elements, so that the vector becomes a list $ ([y_1,id_1],[y_2,id_2],\cdots,[y_m,id_m]) $. Secondly, the list is sorted by $ y_i $ in descending order. Finally, the order of IDs is figured out from the sorted list, and the order decides a ranking list of rules.
\end{itemize}

\subsubsection{The Education of SLDR-DL Systems}
\label{subsubsec:edu}

We use the word ``education" instead of ``training" because the process of optimising an SLDR-system is usually from simple problems to complex problems and requires the interaction between learning and reasoning, and this process is similar to the process of educating a human. In other words, resolution in SLDR-DL is a heuristic search process which can optimise its search strategy via learning. Before learning, it can resolve simple goals, but the resolution of complex goals may fail, because the search space may be huge. After learning in proper ways, the search space can be reduced, so that the complex goals can be resolved successfully. Therefore, the education of an SLDR-DL system usually requires a schedule in which problems are sorted from simplest to hardest. By the schedule, the system tries to resolve simple problems at the beginning, works out resolution records and learns the records. Then the system proceeds to more complex problems and continues learning until all problems are resolved.

\subsubsection{A Prolog Library of Deep Neural Networks}
\label{subsubsec:library}

The SLDR-DL framework also provides a Prolog library which supports essential neural network computations. Specifically, the library now supports:
\begin{itemize}
\item Matrix addition and multiplication.
\item The back-propagation algorithm of feedforward neural networks.
\item The Softmax classifier.
\end{itemize}
Details of the above functions can be found from \cite{Bishop2006Pattern}. To expanded the use of the framework, more functions will be added to the library in the future.

\section{A Practical Guide}
\label{sec:guide}

To build and use an SLDR-DL system, users need to define a rule set, a symbol set and a neural network. These definitions should be coded in Prolog (preferably SWI-Prolog) \cite{DBLP:journals/tplp/WielemakerSTL12}.

\subsection{Defining a Rule Set}
\label{subsec:defineruleset}

\begin{figure}[!t]
\centering
\includegraphics[width=3in]{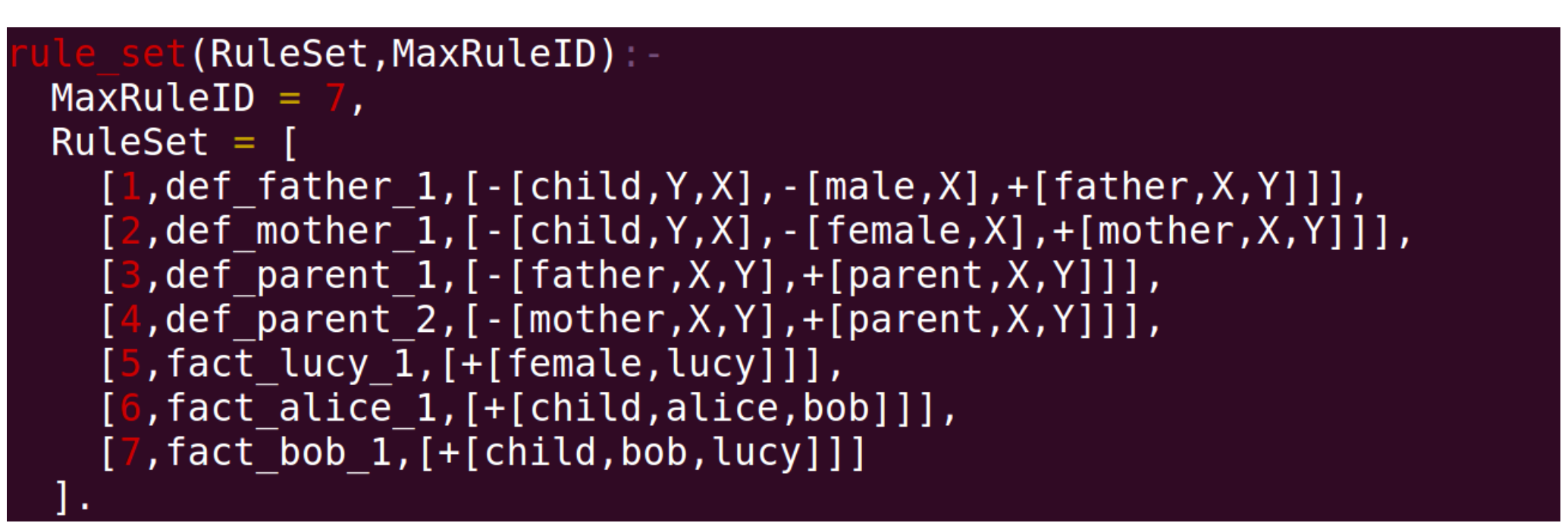}
\caption{Defining a Rule Set.}
\label{fig:001}
\end{figure}

A rule set is defined as a list of rules (definite clauses) written in disjunction form with their unique IDs and names. A rule is defined in the following format:
\begin{equation}
\label{format_rule}
['Rule~ID',~'Rule~Name',~'Disjunction~Form']
\end{equation}
The disjunction form of a rule $ p_1 \land p_2 \land \cdots \land p_n \implies q $ is defined as:
\begin{equation}
\label{format_clause_1}
[-p_1,-p_2,-p_3,\cdots,-p_n,+q]
\end{equation}
where ``$ - $" denotes a negative literal (premise), and ``$ + $" denotes a positive literal (conclusion). In particular, the number of negative literals can be zero, and the rule becomes an assertion $ [+q] $. Figure \ref{fig:001} provides an example of a rule set, where $ MaxRuleID $ is the maximum ID of rules. It is important to note that when defining a rule set, we use the Prolog convention: A symbol is a constant if it is a number, or its first letter is in lower case. A symbol is a variable if its first letter is in lower case. For instance, ``$ [-[child,Y,X],-[male,X],+[father,X,Y]] $" means that for any $ X $ and $ Y $, if $ Y $ is a child of $ X $, and $ X $ is a male, then $ X $ is the father of $ Y $.

\subsection{Defining a Symbol Set}
\label{subsec:definesymbolset}

\begin{figure}[!t]
\centering
\includegraphics[width=3in]{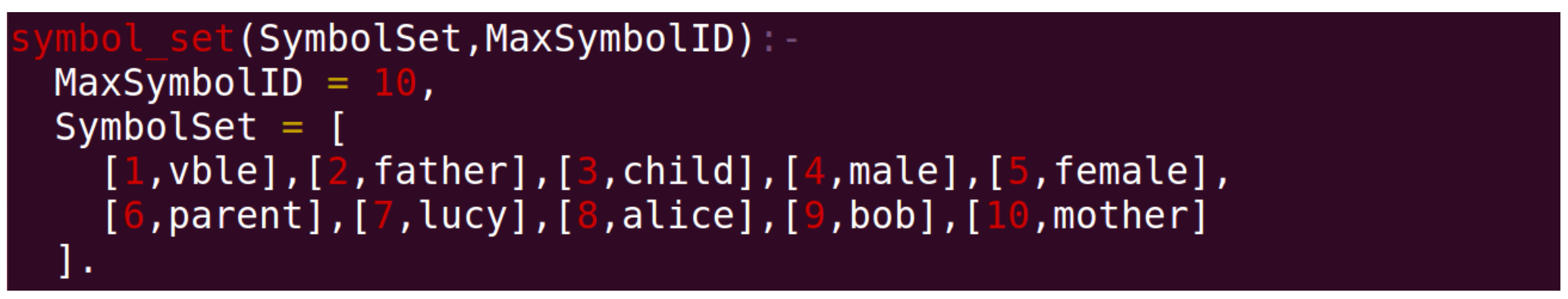}
\caption{Defining a Symbol Set.}
\label{fig:002}
\end{figure}

A symbol set is defined as a list of symbols with their unique IDs. A symbol is defined in the following format:
\begin{equation}
\label{format_rule}
['Symbol~ID',~'Symbol']
\end{equation}
Figure \ref{fig:002} provides an example of a symbol set, where $ MaxSymbolID $ is the maximum ID of symbols.

\subsection{Defining a Neural Network}
\label{subsec:definenn}

\begin{figure}[!t]
\centering
\includegraphics[width=3in]{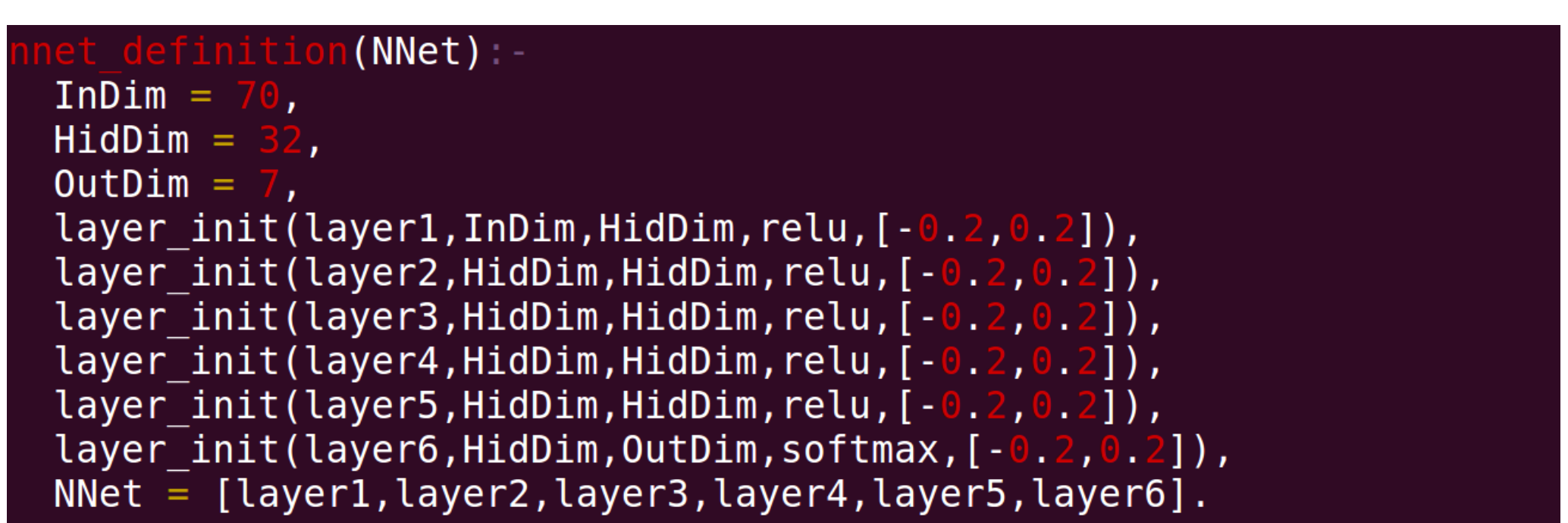}
\caption{Defining a Neural Network.}
\label{fig:003}
\end{figure}

A neural network can be defined as a list of layers:
\begin{equation}
\label{format_dnn}
\begin{aligned}
~[~~&~~ \\
~~~~&'Input~Layer', \\
~~~~&'Hidden~Layer~1', \\
~~~~&'Hidden~Layer~2', \\
~~~~&\cdots, \\
~~~~&'Hidden~Layer~N', \\
~~~~&'Output~Layer' \\
~]~~&~~
\end{aligned}
\end{equation}
Each layer can be initialised via:
\begin{equation}
\label{format_layer}
\begin{aligned}
~lay&er\_init(~~ \\
~~~~&'Layer~Name', \\
~~~~&'Input~Dimension', \\
~~~~&'Output~Dimension', \\
~~~~&'Activation~Type', \\
~~~~&'Scale~of~Randomisation', \\
~)~~&~~
\end{aligned}
\end{equation}
Figure \ref{fig:003} provides an example of the definition of a neural network.

\subsection{Learning and Reasoning}
\label{subsec:learn_reason}

\begin{figure}[!t]
\centering
\includegraphics[width=3in]{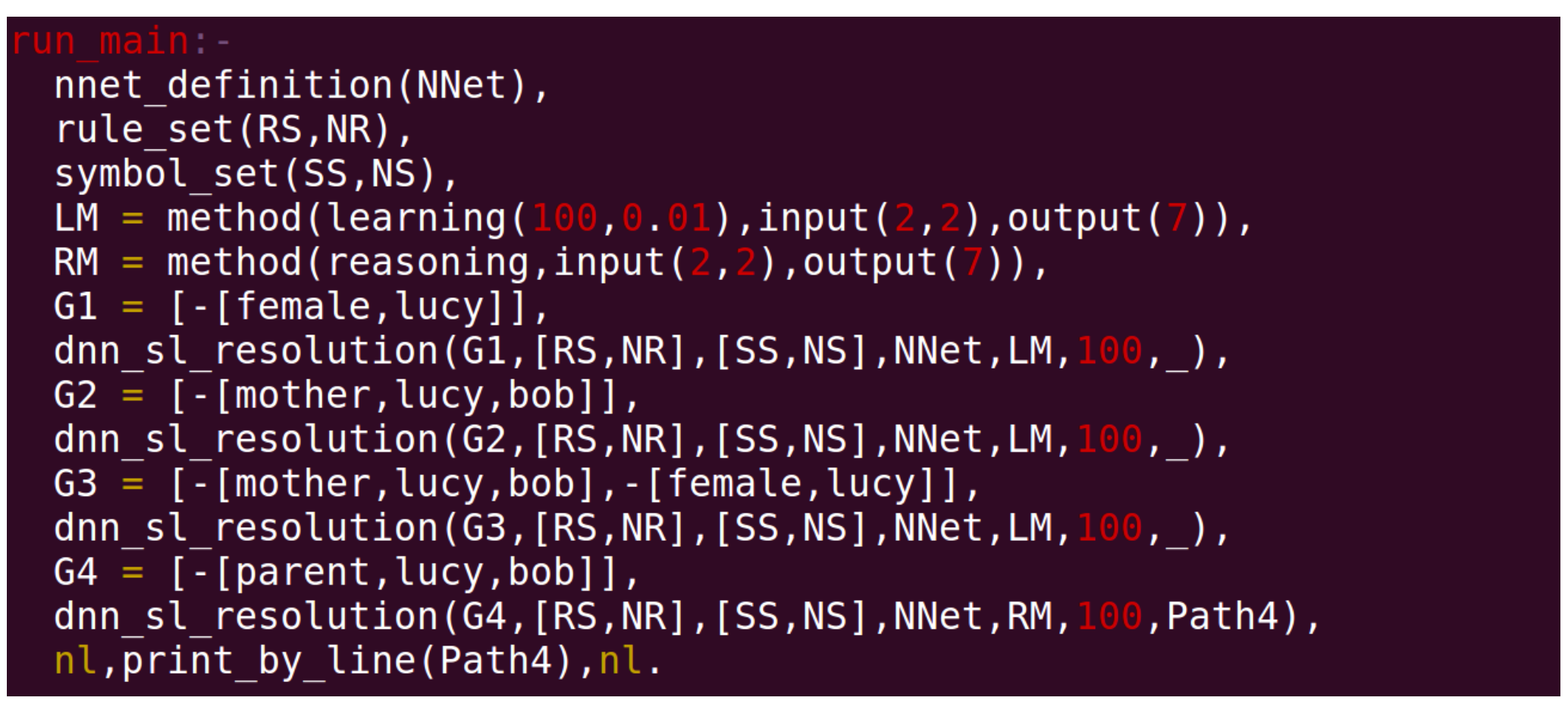}
\caption{Using the Core Function.}
\label{fig:004}
\end{figure}

\begin{figure}[!t]
\centering
\includegraphics[width=3in]{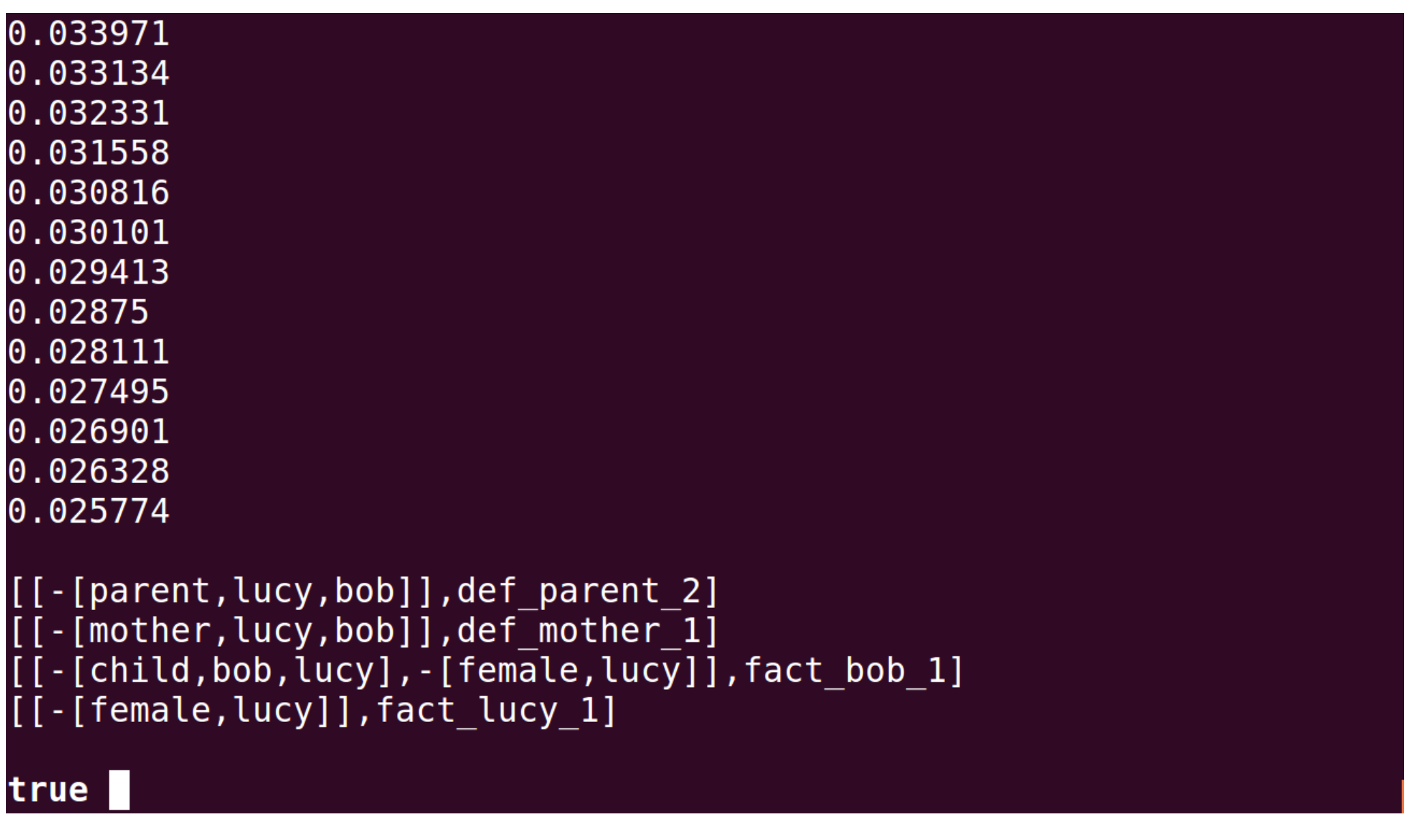}
\caption{A Result of Running.}
\label{fig:005}
\end{figure}

The framework provides a core function named ``$dnn\_sl\_resolution $":
\begin{equation}
\label{dnn_sl_resolution}
\begin{aligned}
~dnn&\_sl\_resolution( \\
~~~~&'Goal', \\
~~~~&'Rule~Set', \\
~~~~&'Symbol~Set', \\
~~~~&'Neural~Network', \\
~~~~&'Method',\\
~~~~&'Search Depth', \\
~~~~&'Result' \\
~)~~&~~
\end{aligned}
\end{equation}
Both learning and reasoning processes are based on the core function. Figure \ref{fig:004} provides an example about how to use the core function, where $ G1 $, $ G2 $, $ G3 $ and $ G_4 $ are goals, ``$ learning(N,R) $" is used to define the number of learning epochs $ N $ and the learning rate $ R $, ``$ input(B,D) $" is used to define the breadth $ B $ and the depth $ D $ of encodings, and ``$ output(Y) $" is used to define the dimension of decodings $ Y $. When running the process, the neural network learns from the resolution processes of $ G1 $, $ G2 $ and $ G3 $ and then tries to resolve $ G4 $. Figure \ref{fig:005} shows a result of running, including a record of cross-entropy losses and a resolution process of $ G4 $.

\section{Summary}
\label{sec:summ}

The SLDR-DL framework enables the interaction between resolution and deep learning. In the framework, users can define logical rules in the form of definite clauses, define neural networks and teach the neural networks to use the logical rules. The neural networks can learn from successful resolution processes and then use learnt experiences to guide new resolution processes. To expand the use of this framework, we will add more functions to it and refine it in the future.

\bibliographystyle{unsrt}
\bibliography{reference}

\end{document}